\documentclass[letterpaper, 10 pt, conference]{ieeeconf}  % Comment this line out if you need a4paper
\usepackage{amsmath}
\usepackage[utf8]{inputenc}
\usepackage{cite}
\usepackage{graphicx}
\usepackage{url}
\usepackage[ruled]{algorithm2e}
\usepackage{amsfonts}
\usepackage{booktabs}
\usepackage{array}
\newcolumntype{P}[1]{>{\centering\arraybackslash}p{#1}}
\newcommand{\CarD}[1]{\mathbf{Card}(#1)}
\newcommand{\N}{\mathcal{N}}
\newcommand{\X}{\mathcal{X}}
\newcommand{\Q}{\mathcal{Q}}
\newcommand{\Z}{\mathcal{Z}}

\newcommand{\norm}[1]{\left\lVert#1\right\rVert}

\newcommand{\pa}[1]{\left(#1\right)}

\newcommand{\SuMjNh}{\sum_{j\in \mathcal{N}_h}}
\newcommand{\SuMiNh}{\sum_{i\in \mathcal{N}_h}}
\newcommand{\SuMiN}{\sum_{i\in \mathcal{N}}}
\newcommand{\SuMkN}{\sum_{k\in \mathcal{N}}}
\newcommand{\SumiN}{\sum_{i=1}^N}
\newcommand{\SuMjm}{\sum_{j=1}^m}
\newcommand{\SuMjpm}{\sum_{j'=1}^m}
\newcommand{\SuMjpneq}{\sum_{j'\neq j^*}}
\newcommand{\Dij}{d_{ij}}
\newcommand{\xb}{\overline{x}}
\newcommand{\Xb}{\overline{X}}
\newcommand{\Rb}{\mathbb{R}}
\newcommand{\Pyx}{p(y_j\mid x_i)}
\newcommand{\Pxy}{p(x_i\mid y_j)}
\newcommand{\Dxy}{d(x_i, y_j)}
\newcommand{\Dyy}{d(y_j, y_{j'})}
\newtheorem{theorem}{Theorem}
\newcommand{\MiNjm}{\min_{j\in\{1,\ldots, m\}}}

\DeclareMathOperator*{\argmin}{arg\,min}
\DeclareMathOperator*{\ArgMinjm}{arg\,min_{j\in\{1,\ldots,m\}}}
\newcommand{\Yjs}{y_{j^*}}
\newcommand{\Yjp}{y_{j'}}
\newcommand{\Pysx}{p(y_{j^*}\mid x_i)}

\newtheorem{proposition}{Proposition}

\IEEEoverridecommandlockouts                              % This command is only needed if 
\title{\LARGE \bf
A Clustering Approach to Edge Controller Placement in Software Defined Networks with Cost Balancing
}

\author{Reza Soleymanifar, Amber Srivastava, Carolyn Beck, Srinivasa Salapaka% <-this % stops a space
\thanks{The authors are with Coordinated Science Laboratory,
        University of Illinois at Urbana-Champaign, 1308 W Main St, Urbana, IL 61801
        {\tt\small $\{$reza2, asrvstv6, beck3, salapaka$\}$@illinois.edu}}%
}

\begin{document}

\maketitle
\thispagestyle{empty}
\pagestyle{empty}

%%%%%%%%%%%%%%%%%%%%%%%%%%%%%%%%%%%%%%%%%%%%%%%%%%%%%%%%%%%%%%%%%%%%%%%%%%%%%%%%
\begin{abstract}
    In this work we introduce two novel deterministic annealing based clustering algorithms to address the problem of Edge Controller Placement (ECP) in wireless edge networks. These networks lie at the core of the fifth generation (5G) wireless systems and beyond. These algorithms, ECP-LL and ECP-LB, address the dominant leader-less and leader-based controller placement topologies and have linear computational complexity in terms of network size, maximum number of clusters and dimensionality of data. Each algorithm tries to place controllers close to edge node clusters and not far away from other controllers to maintain a reasonable balance between synchronization and delay costs. While the ECP problem can be conveniently expressed as a multi-objective mixed integer non-linear program (MINLP), our algorithms outperform state of art MINLP solver, BARON both in terms of accuracy and speed. Our proposed algorithms have the competitive edge of avoiding poor local minima through a Shannon entropy term in the clustering objective function. Most ECP algorithms are highly susceptible to poor local minima and greatly depend on initialization.
    
    \textbf{Keywords}: Clustering, deterministic annealing, 5G networks, software defined networks, wireless edge networks, edge controller placement 
\end{abstract}

%%%%%%%%%%%%%%%%%%%%%%%%%%%%%%%%%%%%%%%%%%%%%%%%%%%%%%%%%%%%%%%%%%%%%%%%%%%%%%%%
\section{Introduction}
Wireless networks are of high importance in modern telecommunication systems as they are efficient, mobile, responsive, accessible, have enhanced guest access and better support expansion of network. In order to enhance these systems, Software-Defined Networks (SDN) have been introduced as an emerging paradigm whose primary advantage is giving developers greater control over the network traffic and administration \cite{Alshamrani2018FaultEnvironments}. Traditionally wireless networks have played both the role of administration and relay of data within the same infrastructure. One of the limitations of this architecture is that modifying these networks requires manually re-configuring nodes of the network to accommodate the new changes. Softwarization is a new trend in wireless communication networks that helps to automate this type of manual work. 

\begin{figure}[htb]
\centering
\includegraphics[width=8cm]{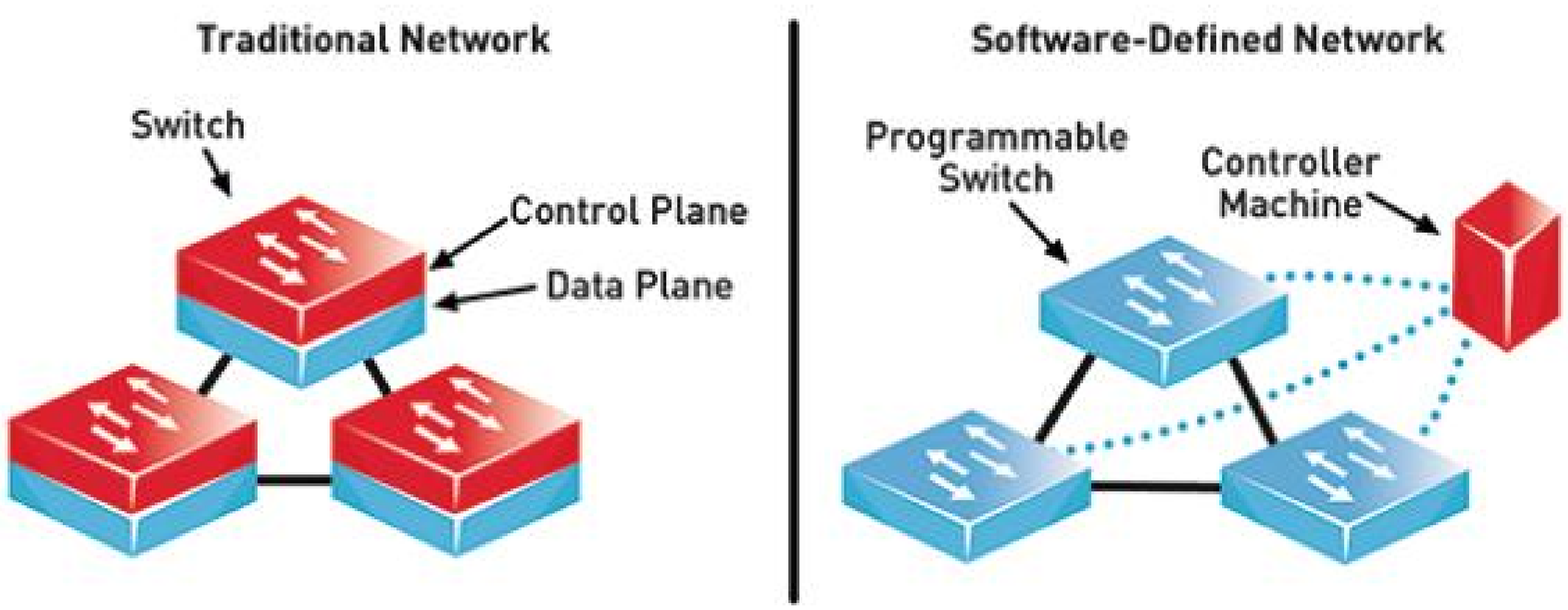}
\caption{Traditional versus Software Defined Networks. Image from \cite{Dungay2016SoftwareBusiness}}
\end{figure}

One of the most studied open research problems, on which SDN itself heavily relies, is the so-called edge controller placement problem (ECP) \cite{Alshamrani2018FaultEnvironments}. Controller placement is one of the most important components of software defined networks \cite{Kuang2018AArchitecture}. This problem was first introduced in \cite{Heller2012TheProblem} and is in general NP-hard \cite{Singh2018ASDN}. Controllers are network nodes which are designated to control other nodes of a network. ECP in a fog/cloud network essentially reduces to determining how many and which nodes in the network need to be designated as the controllers. This placement induces several costs including delays between edge nodes and the controllers they are assigned to, and synchronization delay between the controllers themselves which we refer to as delay and synchronization costs respectively throughout the paper.

 There are several approaches to address ECP problem. Our approach here is based on viewing this problem in a data clustering sense. Many clustering based approaches in literature are hindered by naive initialization and are thus prone to poor local optima. This leads to multiple optimization attempts with varied initializations that increase total computation time needed to find an optimal placement. These approaches are also restricted to a single objective value which prevents the decision maker from simultaneously considering multiple controller placement criteria. 
In this paper, we discuss the use of the deterministic annealing (DA) algorithm, which is tailored to avoid these shortcomings, and introduce algorithms that iteratively minimize the costs associated with ECP. In order to evaluate our algorithms we compare the final costs incurred with those of the MINLP formulation.

We identify the core competences of our algorithms as being (1) scalable and fast, due to linear computational complexity in terms of problem size and number of controllers, (2) high quality in terms of near optimal solutions, (3) initialization independent as we always start with one controller in the mass center of data, (4) excellent at avoiding poor local minima due the to use of a Shannon entropy term in the clustering objective function and (5) able to address a multi-objective scheme. 

The rest of the paper is organized as follows. In Section \ref{sec:lit-review} we overview ECP and SDN related works from recent years. In Section \ref{sec:prob-statement} we concretely define ECP and explain the subtleties of this problem. In Section \ref{sol:approach} we describe our approach to the problem and explain how we adapt the DA to the ECP problem. The reader may refer to Section \ref{sec:results} to see the results of the simulations and finally Section \ref{sec:conc} shows conclusions and avenues for future research.

\section{Literature Review}
\label{sec:lit-review}
The controller placement problem for SDNs was first introduced in \cite{Heller2012TheProblem}.
\cite{Li2018ANetworks} implement the Cuckoo search algorithm for the problem of controller placement in SDNs. \cite{Lu2019ANetworking} identify the main function of SDNs as decoupling the data plane and control plane. They summarize prior research on the controller placement problem into four master categories: latency-oriented, reliability-oriented, cost-based, and multi-objective. They also identify controller-placement as one of the hottest topics in SDN. \cite{Killi2018CooperativeSDN} propose a network partition using a controller placement algorithm based on a mixture of k-means and game theoretic initializations. \cite{Liao2017DensityNetworkings} propose a density based controller placement which uses a clustering algorithm to split the network into multiple sub-networks. \cite{Papa2018DynamicNetwork} consider ECP in the context of satellite networks and study the use-case scenario of SDN-enabled satellite space segments. They design an integer linear program to address this problem. Focusing on reliability aspects of ECP, \cite{Alshamrani2018FaultEnvironments} address maximizing fault-tolerance aspects of controller placement rather than performance. They show sacrificing latency for reliability is generally not a good trade-off except in special cases. 

\cite{Das2018INCEPT:Networks} use a multi-objective optimization model to derive a multi-period roll-out plan for controller placements. A similar problem to ECP, the satellite gateway placement problem, is addressed in detail in \cite{Liu2018JointNetwork}. \cite{ZhiyangSu2015MDCP:Networks} propose a novel scheme to minimize measurement overhead, and formulate the Measurement-aware Distributed Controller Placement (MDCP) problem as a quadratic integer programming problem.

\cite{Jalili2019MultiNetworks} consider an Analytic Hierarchy Process (AHP) to address the multi-criteria controller assignment problem. Apart from latency they also address hop count and link utilization as part of the controller assignment process and use a hybridized ad-hoc genetic algorithm to solve it. \cite{Zhang2018Multi-objectiveNetwork} design a multi-objective controller placement scheme that simultaneously addresses reliability, load balance and low latency. They use the heuristic adaptive bacterial foraging optimization to solve this problem. \cite{Tao2018TheBalancing} derive the specific position of all network controllers by minimizing a linear function of load balance factor and total flow request cost. \cite{Dvir2018WirelessProblem} study the wireless controller placement problem using a multi-objective optimization problem and measure the sensitivity of this placement to variant metrics. 

In this paper we present the first maximum entropy based clustering algorithm to address ECP in wireless edge networks. A tutorial on deterministic annealing for the unfamiliar reader may be found in \cite{Rose1998DeterministicProblems}.
We distinguish our algorithms from previous clustering approaches in that it is the first multi-objective clustering approach to the ECP problem and it does not require initialization.
We found previous algorithms in literature that typically enjoy a fast speed such as Cuckoo search, GA, and other ad-hoc heuristics suffering from susceptibility to poor local optima solutions. On the other hand exact approaches like quadratic integer programming are too slow to be practical for real-case scenarios. Our algorithms address these shortcomings by leveraging their ability to sense and escape poor local minima and at the same time enjoy fast speed due to linear computational complexity in terms of parameters of the problem.

\section{Problem Statement}
\label{sec:prob-statement}
\begin{figure}[!htb]
    \centering
    \includegraphics[width = 8cm]{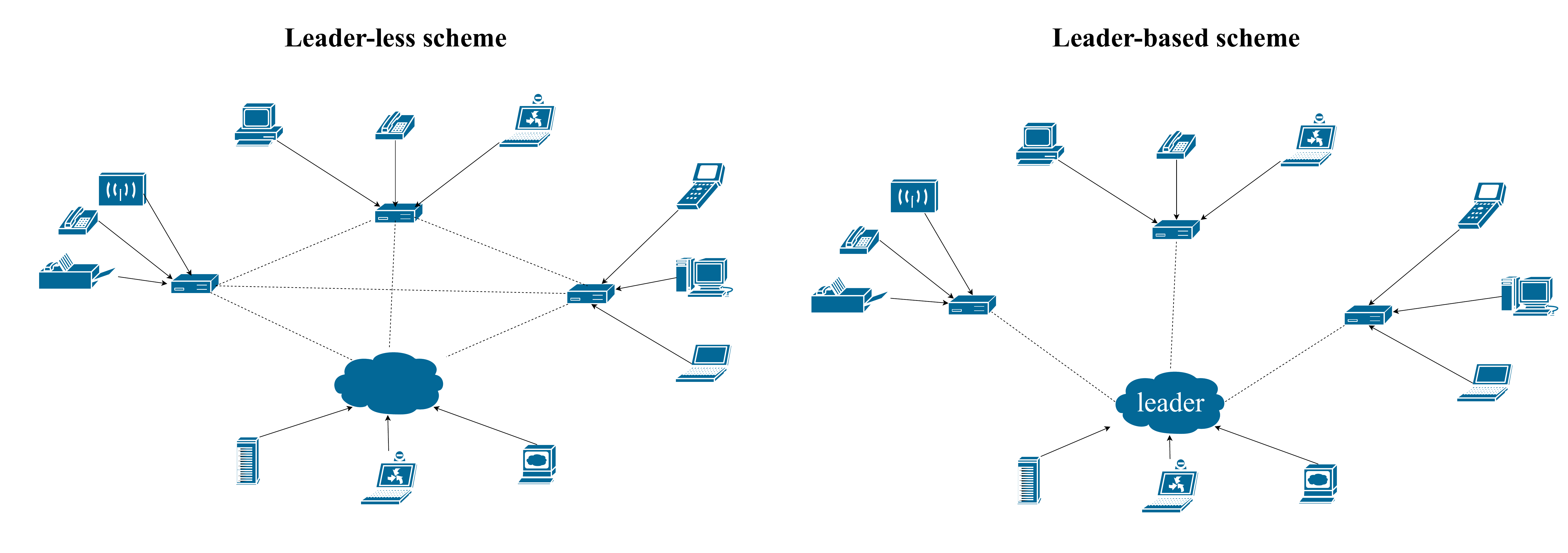}
    \caption{Leader-based versus Leader-less edge controller placement scheme}
    \label{fig:llVslb}
\end{figure}

Wireless networks can be illustrated by a graph as shown in Figure \ref{fig:llVslb}, in which the vertices are the network nodes and the edges represent the communication between them. One or multiple numbers of these vertices can be designated as a controller, where the optimal number and placement of these controllers depends on how far and how close graph vertices are from each other in terms of communication delay associated with graph edges. ECP reduces to finding this optimal assignment of controllers. In this scheme both the nodes and controllers they are assigned to and the controllers themselves constantly communicate data. This means that a scattered placement of controllers may reduce the delay cost but increase the synchronization cost. On the contrary a more compact placement of edge controllers can reduce the synchronization cost while increasing the delay cost between nodes and the controllers. Two dominant schemes for placement of controllers typically considered are leader-less and leader-based \cite{Qin2018SDNNetworks}. The distinction between the two is that in the former all pairs of controllers in the network directly communicate with each other while in the latter controllers only communicate with a leader controller.

To cast this problem as a mathematical program we define $\N$ as the set of all edge nodes with $\CarD{\N} = N$ and
$\N_h$ as the set of edge nodes that can serve as controllers with $\N_h \subseteq \N$. Additionally,
$\X = (x_i \in \Rb^d, i\in \N)$ determines the position of edge nodes in the wireless network. We use $\overline{\X} = (\xb_i \in \{0,1\}, i\in \N_h)$ to represent the controller placement policy. If we choose node $i$ to play the role of a controller then $\xb_i = 1$ otherwise $\xb_i = 0$. Similarly
$\Q = (q_{ij}\in\{0,1\}, i\in \N, j\in \N_h)$ determines the controller assignment policy where $q_{ij}= 1$ if node $i$ is assigned to controller $j$ otherwise $q_{ij} = 0$.
$\Z = (z_j \in \{0, 1\}, j\in \N_h)$ determines the leader assignment policy in the leader-based scheme. $z_j=1$ if controller $j$ is the leader and $z_j = 0$ otherwise. 
$\Dij=d(x_i, x_j)$ encodes the communication delay between nodes $i$ and  $j$ which we assume to be proportional to the squared Euclidean distance, i.e. $\Dij = \|x_i-x_j\|^2_2$.
\subsection{Leader-less Case}
In this setting all controllers communicate not only with edge nodes but also with each other. Thus we incur a controller synchronization cost between \textit{all} pairs of controllers. We can express the optimal assignment as the solution of the following integer program:
\begin{align}
\min_{\Q, \overline{\X}}\quad&\SuMiN\SuMjNh q_{ij}d_{ij}
+\label{eqLL0}
\gamma\sum_{i,j\in\mathcal{N}_h} \xb_i \xb_j 
d_{ij}\SuMkN q_{kj}
\\
\text{s.t.}\quad &\sum_{j\in \N_h} q_{ij} = 1 \quad \forall  i\in \N
\label{eqLL1}\\
&q_{ij} \leq \xb_j\quad
\quad \forall i,j \in \N
\label{eqLL2}\\
&\xb_i \in \{0,1\},  \quad i\in \N_h
\label{eqLL3}\\
&q_{ij}\in\{0,1\},  \quad i\in \N, j\in \N_h,
\label{eqLL4}
\end{align}
The first term in the objective function corresponds to communication delay across all node-controller pairs. The second term shows the synchronization delay between controllers. Note that synchronization delay also depends on how many nodes are assigned to a certain controller. Constraint \eqref{eqLL1} ensures that each edge node is only assigned to one controller; constraint \eqref{eqLL2} ensures node assignments to a controller are only made to designated controller nodes. Parameter $\gamma \geq 0$ shows the relative importance of controller synchronization delay compared to controller-node delay.

\subsection{Leader-based Case}

The leader-based case is similar to the previous one except that controllers synchronize only with the leader. We can express the optimal assignment in this setting as the solution to the following integer program:

\begin{align}
\min_{(\Q, \overline{\X}, \Z)}\quad&\SuMiN\SuMjNh q_{ij}\Dij+
\gamma \SuMiNh\SuMjNh \xb_i z_j 
\pa{
N \Dij
}\label{eqLB0}\\
\text{s.t.}\quad &\sum_{j\in \N_h} q_{ij} = 1\quad \forall i\in \N
\label{eqLB1}\\
&q_{ij} \leq \xb_j\quad
\forall i,j \in \N
\label{eqLB2}\\
&\SuMjNh z_j = 1
\label{eqLB3}\\
&\xb_i \in \{0,1\}, \quad i\in \N_h
\label{eqLB4}\\
&q_{ij}\in\{0,1\}, \quad i\in \N, j\in \N_h\label{eqLB5}.
\end{align}

Constraint \eqref{eqLB3} ensures that there is always exactly one leader controller in the leader-based setting. Both leader-less and leader-based cases are NP-hard nonlinear combinatorial problems with no guarantees for finding a global optimum solution \cite{Singh2018ASDN}. 

\section{Solution Approach}
\label{sol:approach}
 We assume that the delay and synchronization costs are all equivalent to the squared Euclidean distance between the network nodes. This is not far-fetched since according to \cite{Qin2018SDNNetworks} these costs are proportional to if not determined by the Euclidean distances. We further assume that geospatial coordinates\footnote{Here, assumed to be a two dimensional or three dimensional vector representing the location of each edge node.} of the nodes are provided to us instead of the mutual delays between network nodes.
 
 In the deterministic annealing clustering setting, the expected \textit{distortion}~\footnote{Distortion is an average weighted distance term, between nodes and centroids, that serves as our basic cost function.} can be defined as 
 \[D  = \SumiN p(x_i) \SuMjm \Pyx D(x_i, y_j).\]
 $X = \{x_i\}_{i=1}^N$ are the data points and $Y= \{y_j\}_{j=1}^m$ are cluster centroids, or edge controller locations, to be determined. $p(y_j\mid x_i)$ is called the association probability\footnote{The weighting indicating that a node belongs to a particular centroid. For each node the sum of these associations over all centroids must equal one.} of point $x_i$ with centroid $y_j$ and $D(x_i, y_j)$ is the distortion measure which is typically chosen to be the squared Euclidean distance. We interpret $p(x_i)$ as the relative importance given to $i$th node and assume, if not otherwise indicated that $p(x_i)=\frac{1}{N}$. System entropy can be defined as $H = -\SumiN p(x_i)\SuMjm \Pyx\log \Pyx$.
 We also define the system free energy as $F = D - T H$ where $T$ is the system's so-called temperature.\footnote{A coefficient scaling the entropy term which indicates how important the entropy term is compared to the distortion term. We typically reduce this coefficient from a high value to a value close to zero.} Note that $F$ can be viewed as the Lagrangian for the primary objective of minimizing $D$, with $T$ being the Lagrange multiplier. The central iteration of DA can be summarized as sequentially optimizing $F$ with respect to the free parameters, i.e. association probabilities and centroid locations.

\subsection{Leader-less Case}
For the purpose of adapting the DA clustering to the leader-less ECP problem we define the distortion measure as $D(x_i, y_j) = d(x_i, y_j) + \gamma \SuMjpm d(y_j, y_{j'})$. This means the distortion between edge node $x_i$ and controller $y_j$ not only depends on the communication delay between these two nodes but also depends on how far the $y_j$ is placed from other controllers $y_{j'}$.
 
 In order to observe the relation to integer program \eqref{eqLL0}-\eqref{eqLL4} notice we can write total distortion as

\begin{align}
 D= &\SumiN\SuMjm \Pyx \Dxy
\\+ \gamma&\SuMjpm\SuMjm
\pa{
d(y_j, y_{j'})\SumiN \Pyx
}\nonumber
\end{align}

This is objective function \eqref{eqLL0} with hard assignments $q_{ij}$ replaced by the soft association probabilities. As described earlier we define the system's free energy as $ F= D - T H$. Setting partial derivatives of the free energy term with respect to association probabilities to zero and solving, yields solution:
 
\[
\Pyx
=
\frac{\exp\pa{
-\frac{D(x_i, y_j)}{T}
}}
{Z_i},\qquad
Z_i = \SuMjm \Pyx
\]
Thus association probabilities have the celebrated Boltzmann distribution. Similarly setting derivatives with respect to the centroids $y_j$ to zero leads to the following linear systems of equations:
 
\begin{equation}
\eta y_j -\gamma \sum_{j'\neq j} y_{j'} = C_j,\quad
j= 1, \ldots m
\label{eq:LSE}
\end{equation}
 where $\eta = \gamma(m-1) +1$ and $C_j = \SumiN \Pxy x_i$. We may compute $\Pxy$ using Bayes' rule. This gives us a linear system of $md$ variables and $md$ equations with $m$ and $d$ being respectively the number of centroids and the  dimensionality of data. It is essential for the convergence of our clustering algorithm that this linear system of equations always has a solution.
 
 \begin{proposition}
 \label{th:has-sol}
 
 Given the linear system of equations in \eqref{eq:LSE} with $\eta$ and $C_j$ defined as above, if $\gamma \neq \frac{1}{n-m}, \frac{1}{n-2m}$ then there always exists a unique solution $\{y_j \}_{j=1}^m$, where the coefficient matrix associated with the system of the equations is non-degenerate with determinant $\pa{\frac{(\gamma m +1)^m\pa{\gamma(n-m) -1)}}{\gamma(n-2m) -1}}^d$.
 \end{proposition}

See Appendix for a proof. The resulting DA clustering algorithm for the this case is given in Algorithm \ref{alg:LL}.
 
\begin{algorithm}
\small
\label{alg:LL}
 Set max \# of clusters $K_{max}$ and min temperature $T_{min}$\;
 
 Initialize: $T\rightarrow \inf, K=1, y_1 = \SumiN x_i p(x_i)$\;
 \While{Convergence test}{
  Update: 
  \[\Pyx\longleftarrow
  \exp\pa{
-\frac{\Dxy + \gamma \SuMjpm\Dyy}{T}
}/Z_i\;\]
\normalsize
Solve: 
\[\eta y_j^{new} -\gamma\SuMjpneq y_{j'}^{new} = \SumiN \Pxy x_i,\quad
j= 1, \ldots m\]
Update: $y_j \longleftarrow y_j^{new}\;
j= 1, \ldots m$\;
  \eIf{$T\leq T_{min}$}{
   break\;
   }{
   Cooling Step: $T \longleftarrow \alpha T (\alpha < 1)$\;
   Generate small random vector $\epsilon$ 
   
   Replace $y_j$ with $y_j+\epsilon$ and $y_j - \epsilon$\;
  }
 }
Perform last step iteration for $T=0$\;
$y_j \longleftarrow \argmin_{x_i\in \N_h}\Dxy$\;
\caption{ECP-LL}
\end{algorithm}
\normalsize
For the convergence test we stop at iteration $\tau$ if $\norm{F_\tau - F_{\tau - 1}} < \delta$ for some predetermined tolerance level $\delta$. In the last line of Algorithm \ref{alg:LL} we designate the closest valid node to each centroid as a controller.

The iteration complexity for this algorithm depends on (a) calculation of mutual squared Euclidean distances between $x_i,\; y_j$ for $i\in\{1,\ldots,N\}, j\in\{1,\ldots,m\}$, (b) similar calculation of mutual distances between centroids, (c) calculation of association probabilities and (d) solving the linear system of equations. The complexities for these operations are respectively, $O(N K_{max} d)$, $O(K_{max}^2 d)$, $O(K_{max} N)$ and $O(K_{max}^3 d^3)$. For large $N$ these terms are dominated by $O(N  K_{max}  d)$, thus for a maximum number of iterations $\tau$ the algorithmic computational complexity for the leader-less case is $O(\tau N  K_{max}  d)$ which is linear in data size, maximum number of clusters and dimensionality of data.
\subsection{Leader-based case}

In order to adapt DA to the leader-based ECP problem we define an appropriate distortion measure by:
\begin{equation}
    D(x_i, y_j) = 
    \Dxy + \gamma\MiNjm \SuMjpm \Dyy
\end{equation}
Similarly we can consider the weighted  total distortion as:
\begin{equation}
    D = \SumiN\SuMjm \Pyx \Dxy +\gamma\MiNjm \SuMjpm N \Dyy
\end{equation}
In order to observe its relation to MINLP objective function, notice \eqref{eqLB0} is equivalent to the following objective function:
\[
\min_{(\Q, \Xb)}\quad\SuMiN\SuMjNh q_{ij}\Dij
+\gamma\MiNjm\SuMiNh \xb_i  
\pa{
N \Dij
}\label{eqLB20}
\]
To establish this equivalence, we used the relationship that for $W = \{w_i\}_{i=1}^m$ and $S = \{W\in \Rb^n_+ \mid \sum_{i=1}^m w_i = 1\}$ then $\min_{\Z \in S}
\SuMjm z_j \alpha_j =
\min_{j\in \{1,\ldots,m\}} \alpha_j$.

We define the system's free energy similarly to the previous case. Setting the gradient with respect to the association probabilities to zero, yields solution:

\[
\Pyx=
\frac{\exp\pa{
-\frac{\Dxy}{T}
}}
{Z_i},\qquad
Z_i = \SuMjm \Pyx
\]
Denote $\displaystyle j^* = \ArgMinjm \SuMjpm \Dyy$ as the index of the leader centroid and set gradient with respect to $y_j$ to zero to yield the centroid update rules:
\begin{align}
&y_j = \frac
{\gamma N y_{j^*}+\SumiN \Pyx x_i}{\gamma N + \SumiN  \Pyx}
\quad y_j \neq \Yjs\label{eqLBR0}\\
&\Yjs = \frac
{\gamma N \SuMjpneq \Yjp+\SumiN \Pysx x_i}{(m-1)\gamma N + \SumiN  \Pysx}
\label{eqLBR1}
\end{align}

We can compute values of $y_j$ and $\Yjs$ by substituting \eqref{eqLBR1} in \eqref{eqLBR0}. The resulting DA clustering algorithm for the leader-based case can be found in Algorithm \ref{alg:LB}.

\begin{algorithm}
\label{alg:LB}
\small
 Set limits: max \# of clusters $K_{max}$ and minimum temperature $T_{min}$\;
 
 Initialize: $T\rightarrow \inf, K=1, y_1 = \SumiN x_i p(x_i)$\;
 
 \While{Convergence test}{
  Update: $\Pyx\longleftarrow
  \exp\pa{
-\frac{\Dxy}{T}
}/Z_i$\;
Solve: $\displaystyle j^* = \ArgMinjm \SuMjpm \Dyy$\;
Solve: 
\begin{align*}&y_j^{new} = \frac
{\gamma N y_{j^*}+\SumiN \Pyx x_i}{\gamma N + \SumiN  \Pyx}\quad y_j \neq \Yjs
\;\;\\
&\Yjs^{new} = \frac
{\gamma N \SuMjpneq \Yjp+\SumiN \Pysx x_i}{(m-1)\gamma N + \SumiN  \Pysx}
\end{align*}
\normalsize

Update: $y_j \longleftarrow y_j^{new}\;
j= 1, \ldots m$\;
  \eIf{$T\leq T_{min}$}{
   break\;
   }{
   Cooling Step: $T \longleftarrow \alpha T (\alpha < 1)$\;
   Generate small random vector $\epsilon$ 
   
   Replace $y_j$ with $y_j+\epsilon$ and $y_j - \epsilon$\; 
  }
 }
Perform last step iteration for $T=0$\;
$y_j \longleftarrow \argmin_{x_i\in \N_h}\Dxy$\;
\caption{ECP-LB}
\end{algorithm}

The computational complexity for the leader-based algorithm is similar to the previous one, except for the centroid calculation step in which we no longer have to compute a linear system of equations. The computational complexity is $O(N  K_{max}  d) + O(K_{max}^2  d) + O((N + K_{max})  d) + O(N  K_{max})$. For a maximum of $\tau$ iterations and large $N$ this is again dominated by $O(\tau  N  K_{max}  d)$.

\subsection{Phase Transformation}
Our proposed algorithms undergo {\em phase transition} phenomenon analogous to the DA clustering algorithm \cite{Rose1998DeterministicProblems}. More specifically, the algorithms illustrated in Section \ref{sec:prob-statement} begin with allocating a single centroid
$y_j=\SuMiN p(x_i)x_i$ at $T=T_{\max} (\rightarrow \infty)$. As $T$ is gradually decreased, there is no perceptible change in solution till a critical value of temperature $T=T_{cr1}$ is reached where the number of centroids increases. Again as $T$ decreases further, we observe no perceptible change in the solution, till another critical temperature $T = T_{cr2}$ value is achieved where the number of centroids once again increases. In fact, the insensitivity of the solution to $T$ between two consecutive critical values allows us to geometrically anneal the temperature in Algorithms \ref{alg:LL} and \ref{alg:LB} which makes them computationally efficient~\cite{Sharma2012Entropy-basedProblems}. We use the second order necessary condition for optimality to determine the explicit values $T_{cr}$'s where the phase transition occurs. In particular, at $T=T_{\max}$, $F$ is a convex function with its global minimum $Y$ satisfying the second order optimality condition $\frac{\partial F^2}{\partial^2 Y}>0$. As $T$ gradually decreases there occurs an instance $T=T_{cr1}$ where the Hessian $\frac{\partial F^2}{\partial^2 Y}$ loses rank and the number of centroids increases. The following theorem characterizes the temperature $T$ values where phase transitions occur in the case of ECP-LL algorithm.
\begin{theorem}\label{thm: theorem2}
The critical value $T_{cr}$ of the temperature for a given set of centroids $\{y_j\}_{j=1}^m$ and corresponding association weights $\{p(y_j|x_i)\}$ in the ECP-LL algorithm is such that
\begin{small}
\begin{align}\label{eq: phaseTran}
\text{det}\Big[\sum_{i=1}^N p(x_i) \Big(\Lambda_i(1+m\gamma)+\gamma I - 2\gamma \Gamma^T_iE-2\frac{1}{T_{cr}}\Theta_i\Big)\Big] = 0,
\end{align}
\end{small}
where $\Lambda_i,\Gamma_i,E$ and $\Theta_i\in\mathbb{R}^{md\times md}$ are known in terms of $\{x_i\}_{i=1}^N,\{y_j\}_{j=1}^m$ and $\{p(y_j|x_i)\}$, $I$ is an $md\times md$ identity matrix.
\end{theorem}

Similarly, the critical temperature values for the ECP-LB algorithm are determinable. See Appendix for a proof.

\section{Results}
\label{sec:results}
In order to evaluate the performance of these algorithms we compare their final costs with the integer programs \eqref{eqLL0}-\eqref{eqLL4} and \eqref{eqLB0}-\eqref{eqLB5}. We use the state-of-the-art MINLP solver BARON to draw this comparison. We used Gaussian distribution to generate our data with $K$ as the number of Gaussian clusters within the data. During the implementation we perform a grid search over the hyper-parameter space of $K_{max}$ to find its optimum value.

\begin{figure*}[!htb]
\centering
\begin{tabular}{ccc}
\includegraphics[width=55mm]{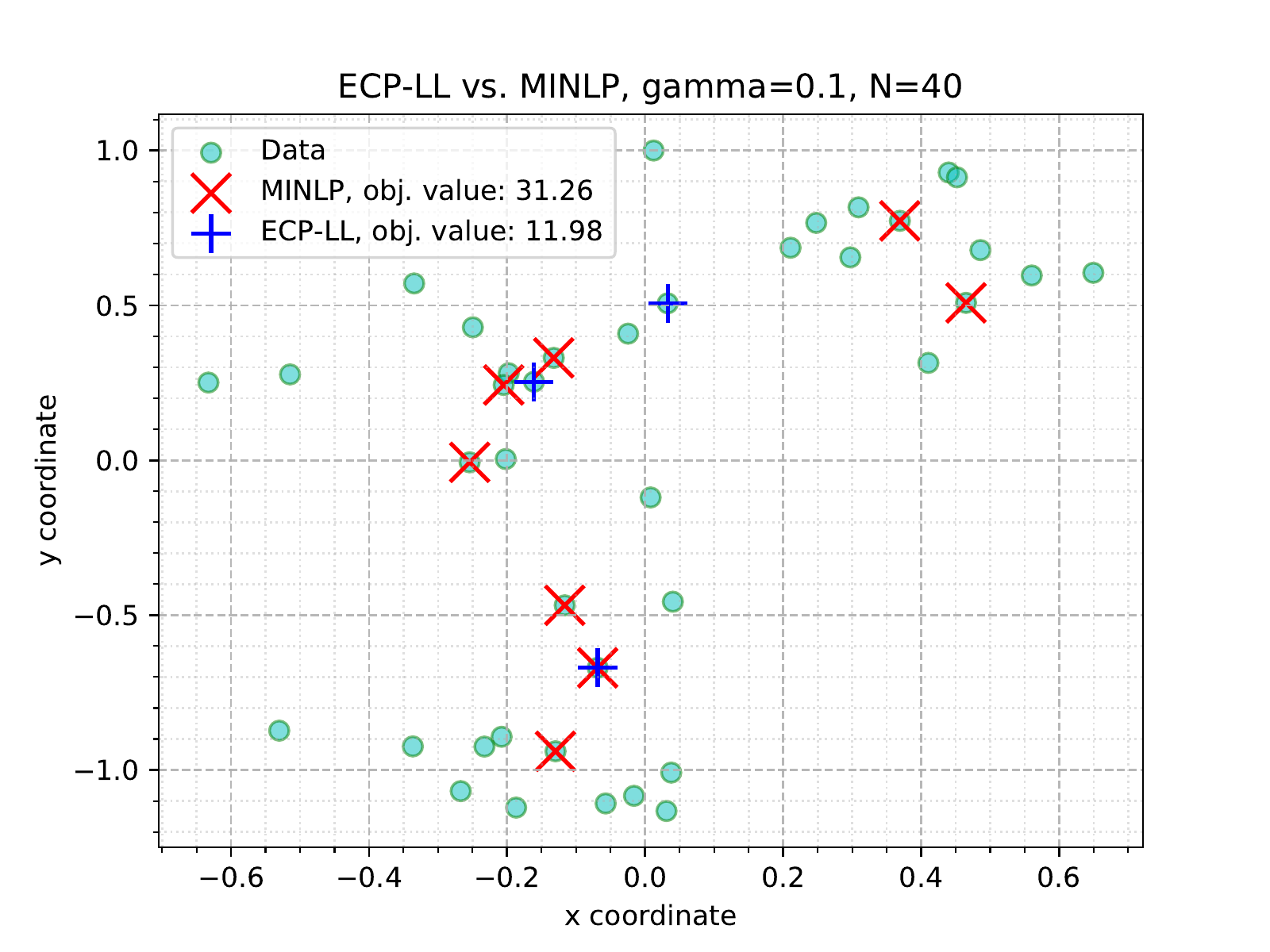}&
\includegraphics[width=55mm]{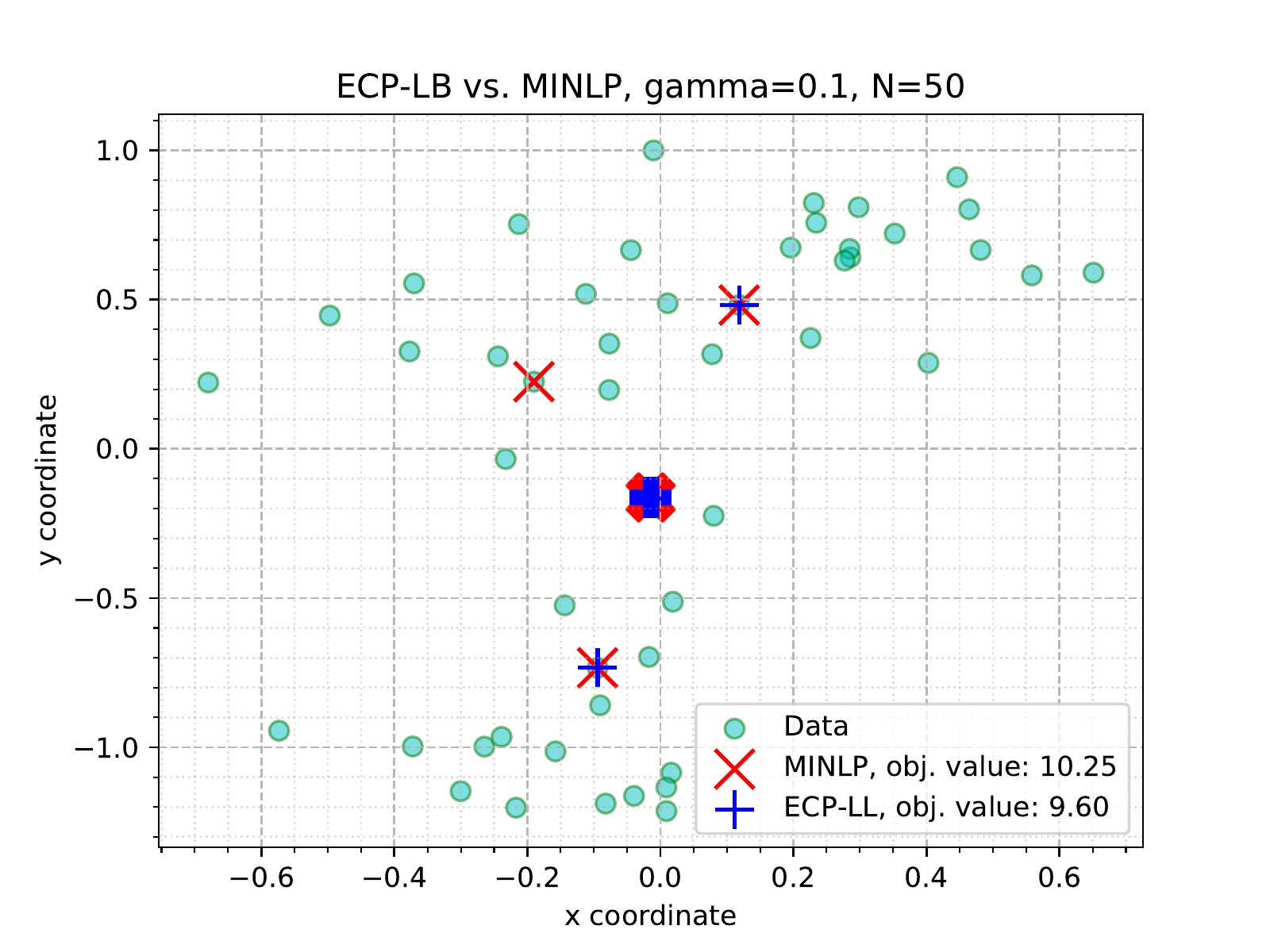}
&
\includegraphics[width=55mm]{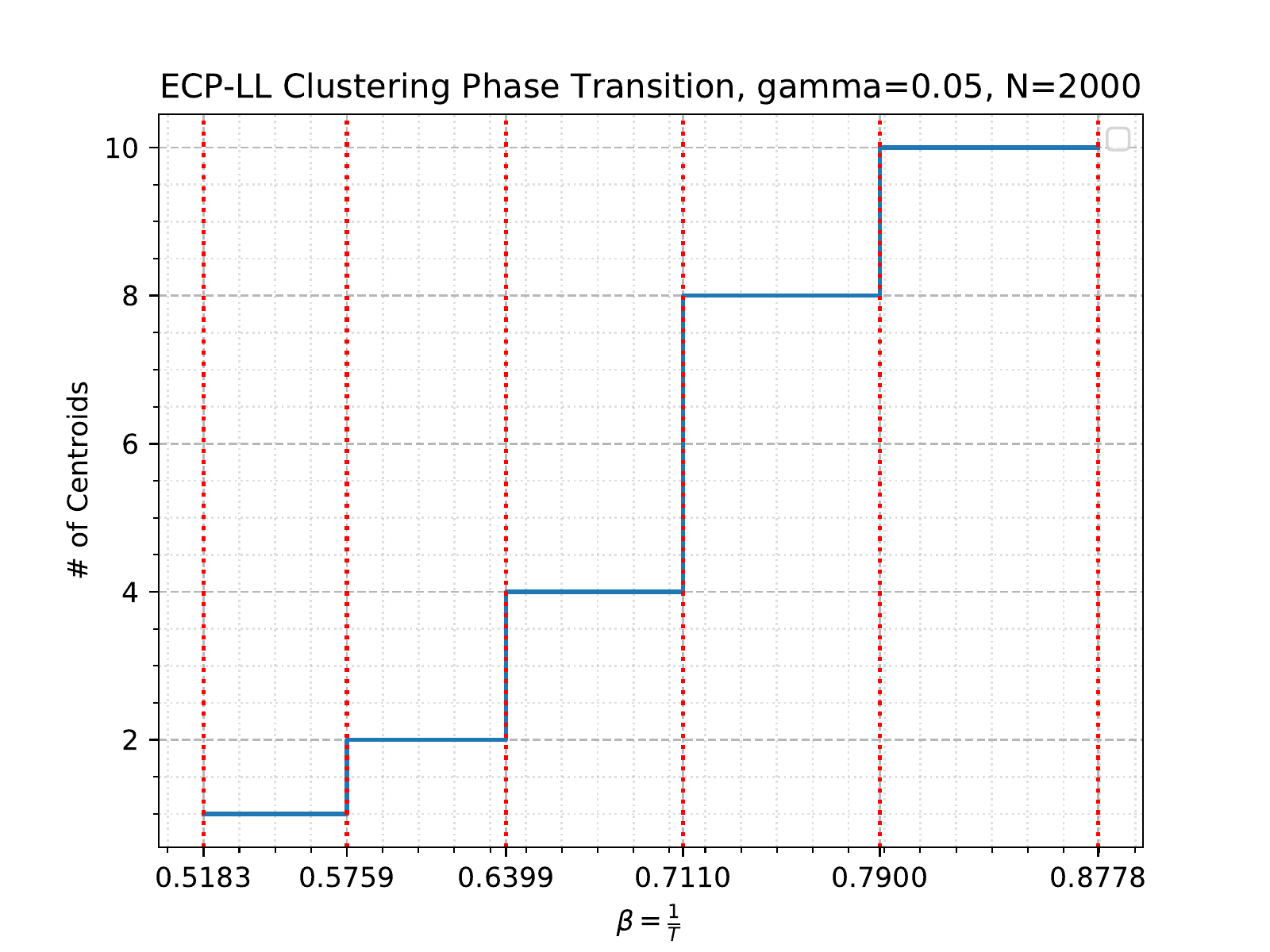}\\
(a) &(b) &(c)
\end{tabular}
\caption{(a) ECP-LL vs. MINLP (b) ECP-LB vs. MINLP (c) Phase transition phenomenon}
\label{fig:clusteringVsMINLP}
\end{figure*}

Superior performance of ECP DA-based clustering algorithms can be observed even in small problem instances like in Figure \ref{fig:clusteringVsMINLP} (a) and (b). While BARON is stuck in a poor local optimum with an excessive number of controllers, ECP-LL has managed to achieve a considerably lower objective value with fewer controller placements.

\begin{figure*}[!htb]
\centering
\begin{tabular}{P{4cm}P{4cm}P{4cm}P{4cm}}
\includegraphics[width=49mm]{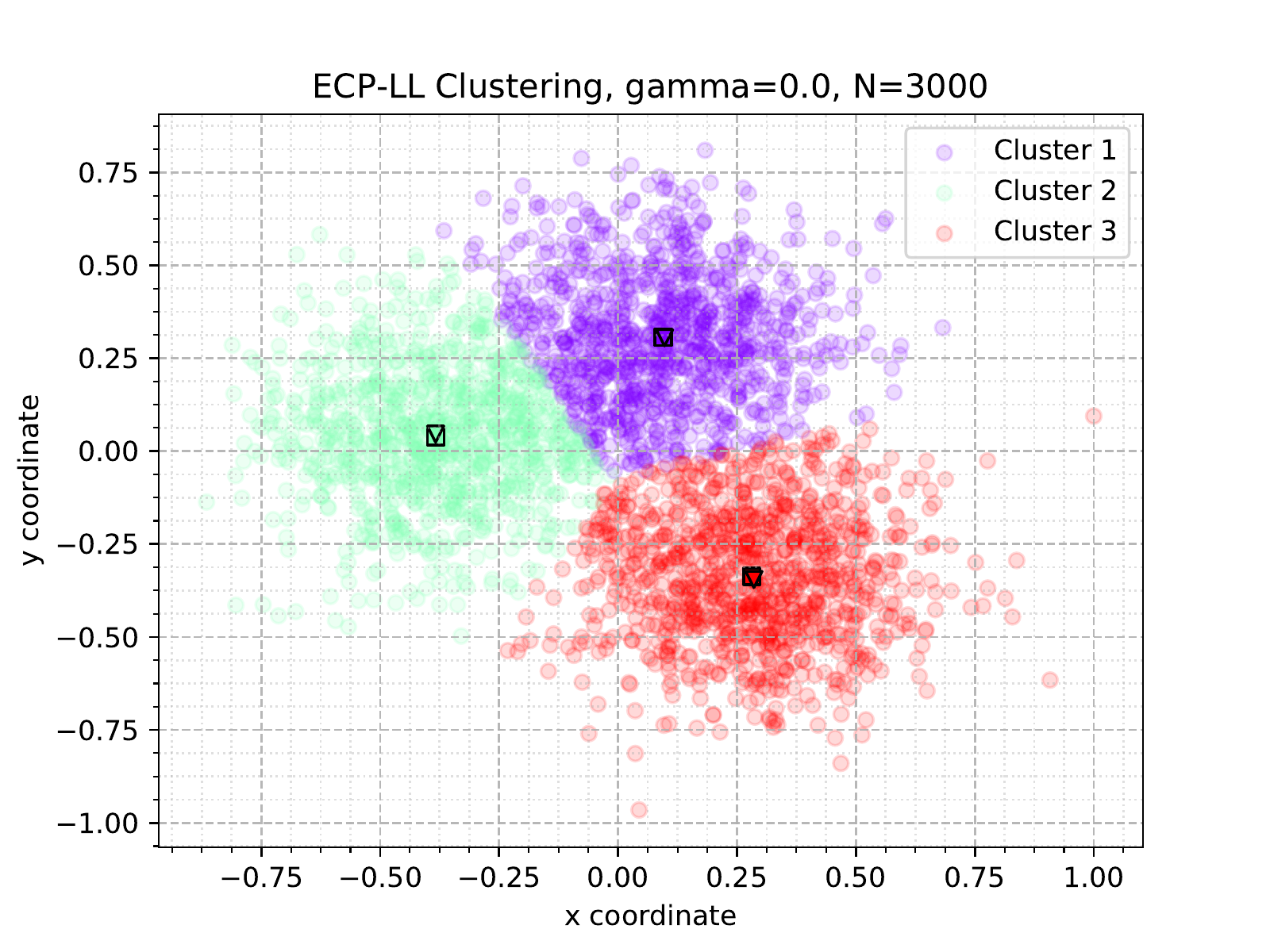}
 &\includegraphics[width=49mm]{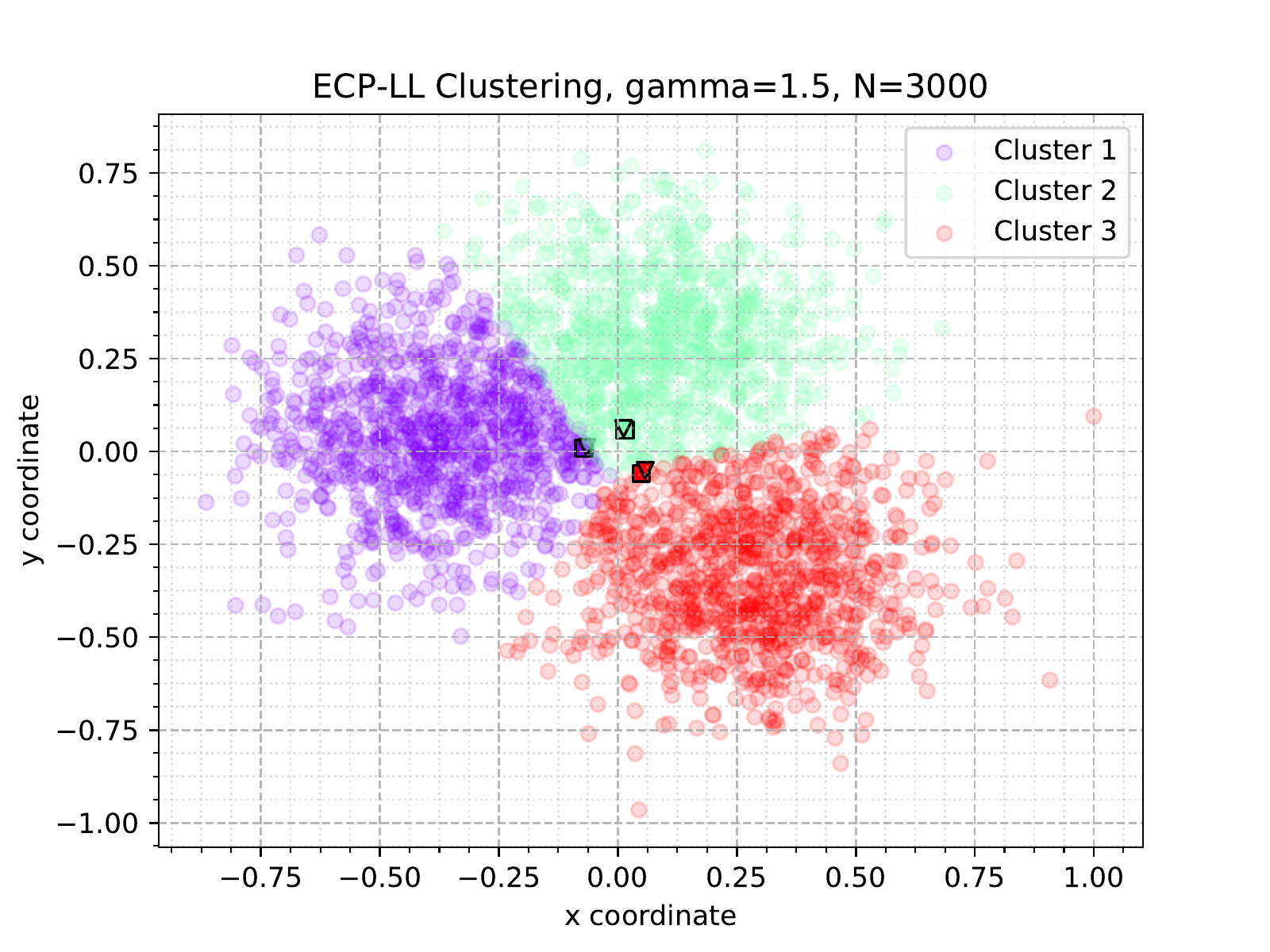}
&\includegraphics[width=49mm]{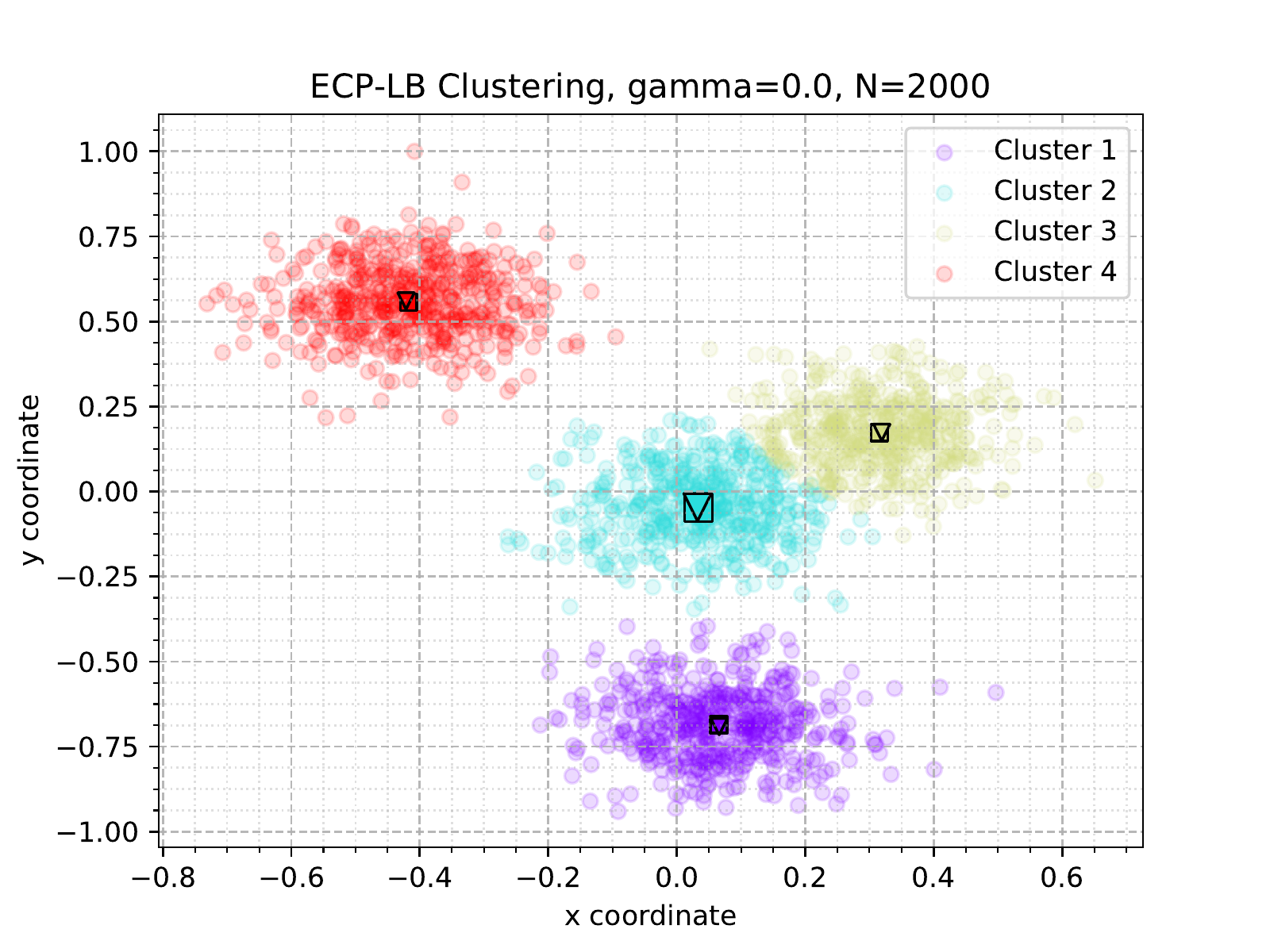}
 &\includegraphics[width=49mm]{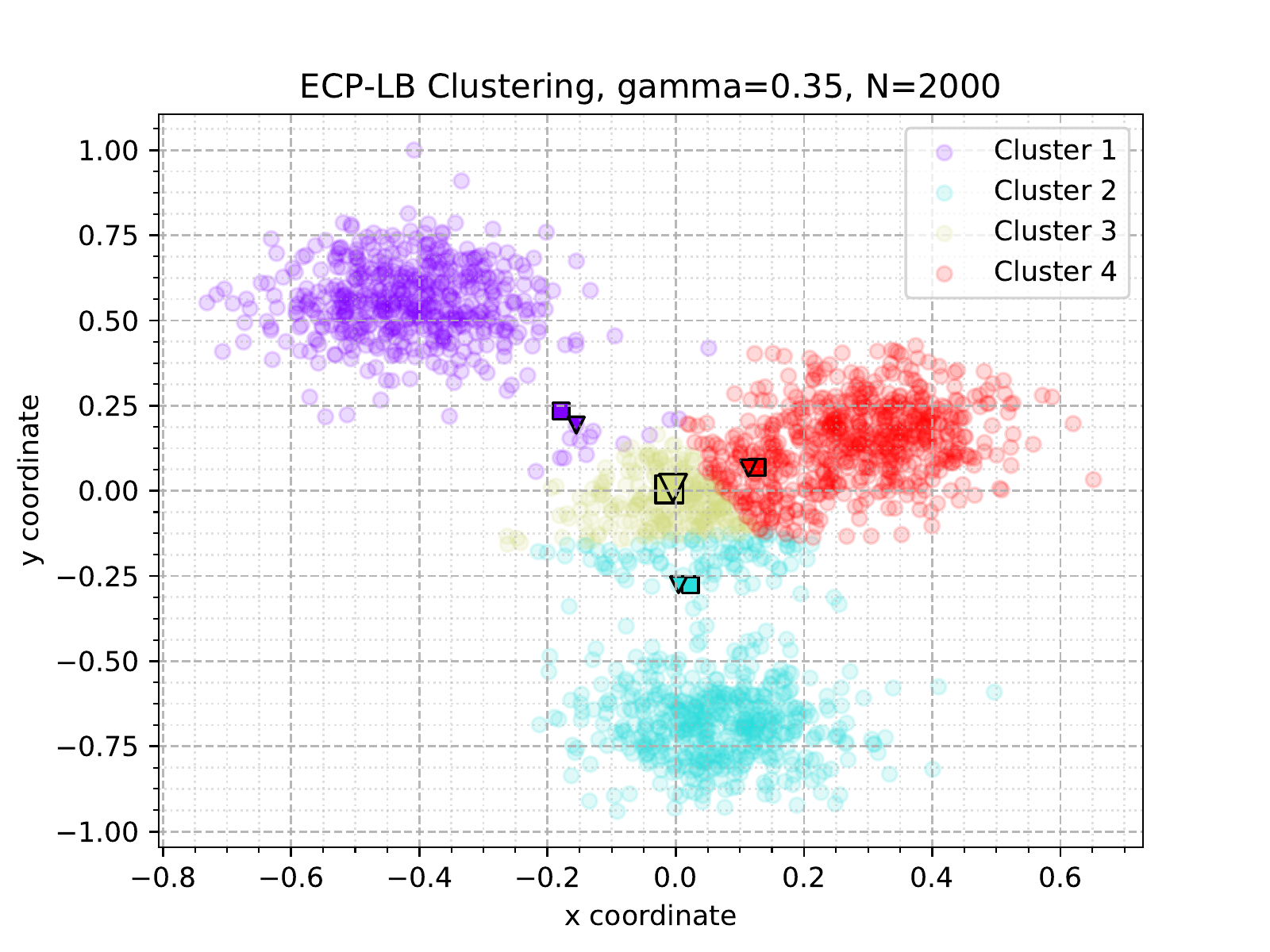}
\end{tabular}
\caption{Controller placement sensitivity to parameter $\gamma$}
\label{fig:LL_sensitivity}
\end{figure*}

% \begin{figure}[!htb]
% \centering
% \includegraphics[width=49mm]{PhaseTrans.pdf}
% \caption{Phase transition phenomenon}
% \label{fig:phase_trans}
% \end{figure}

In Figure \ref{fig:LL_sensitivity}, an immediate result of avoiding controller synchronization cost contributes increasingly to the objective function as $\gamma$ increases.

\begin{figure*}[!htb]
\centering
\begin{tabular}{P{4cm}P{4cm}P{4cm}P{4cm}}
\includegraphics[width=49mm]{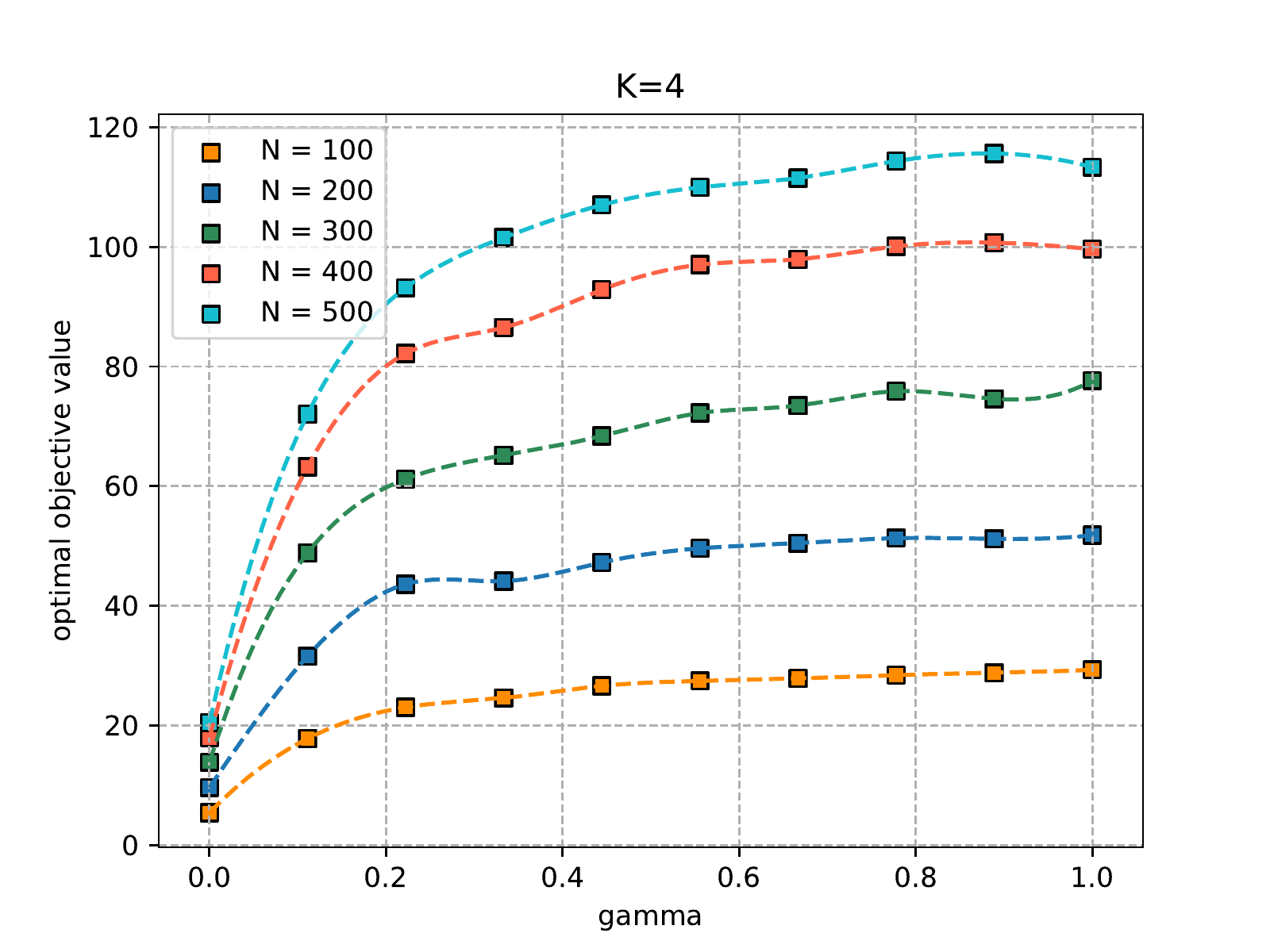}
&\includegraphics[width=49mm]{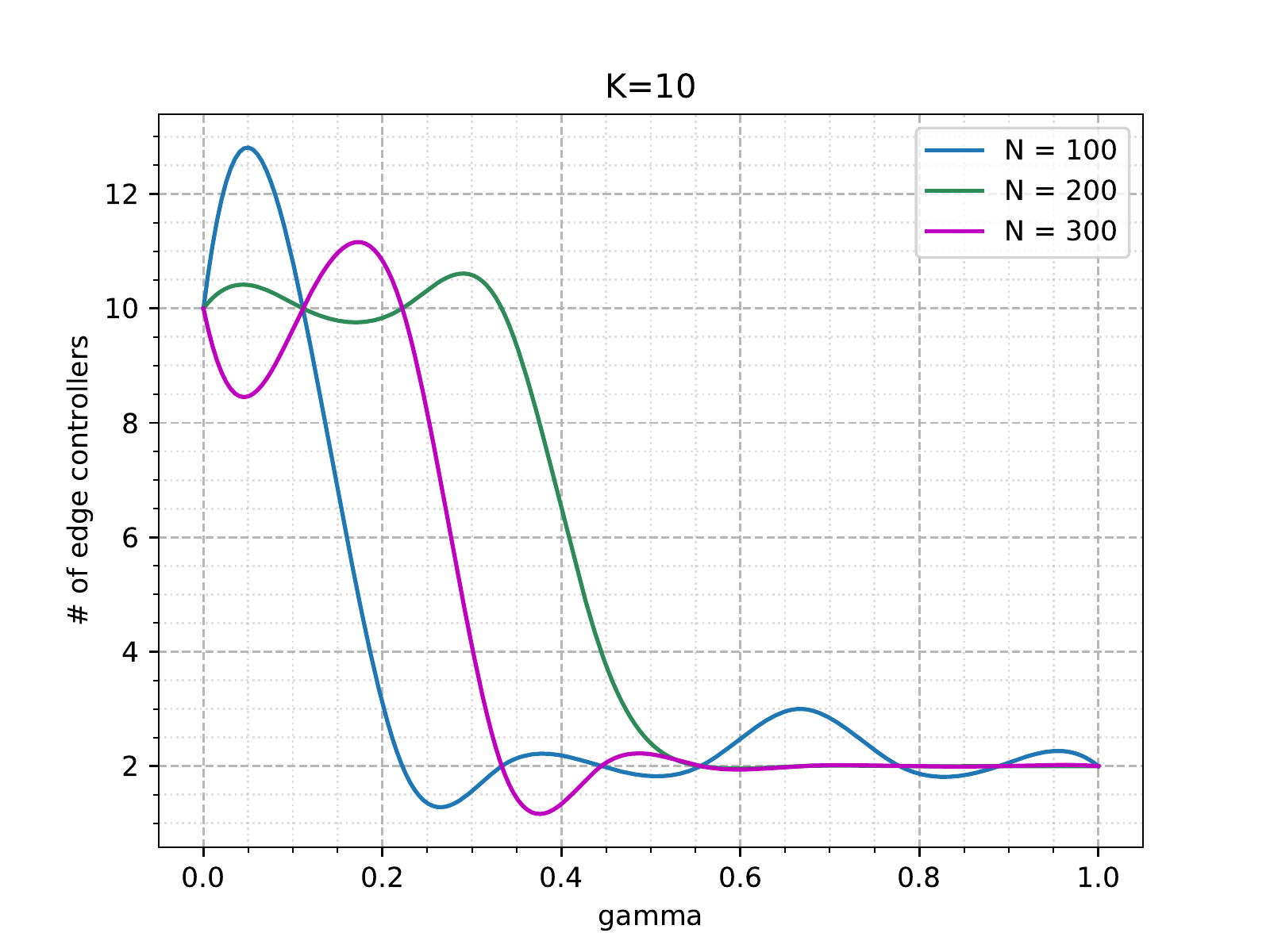}
&\includegraphics[width=49mm]{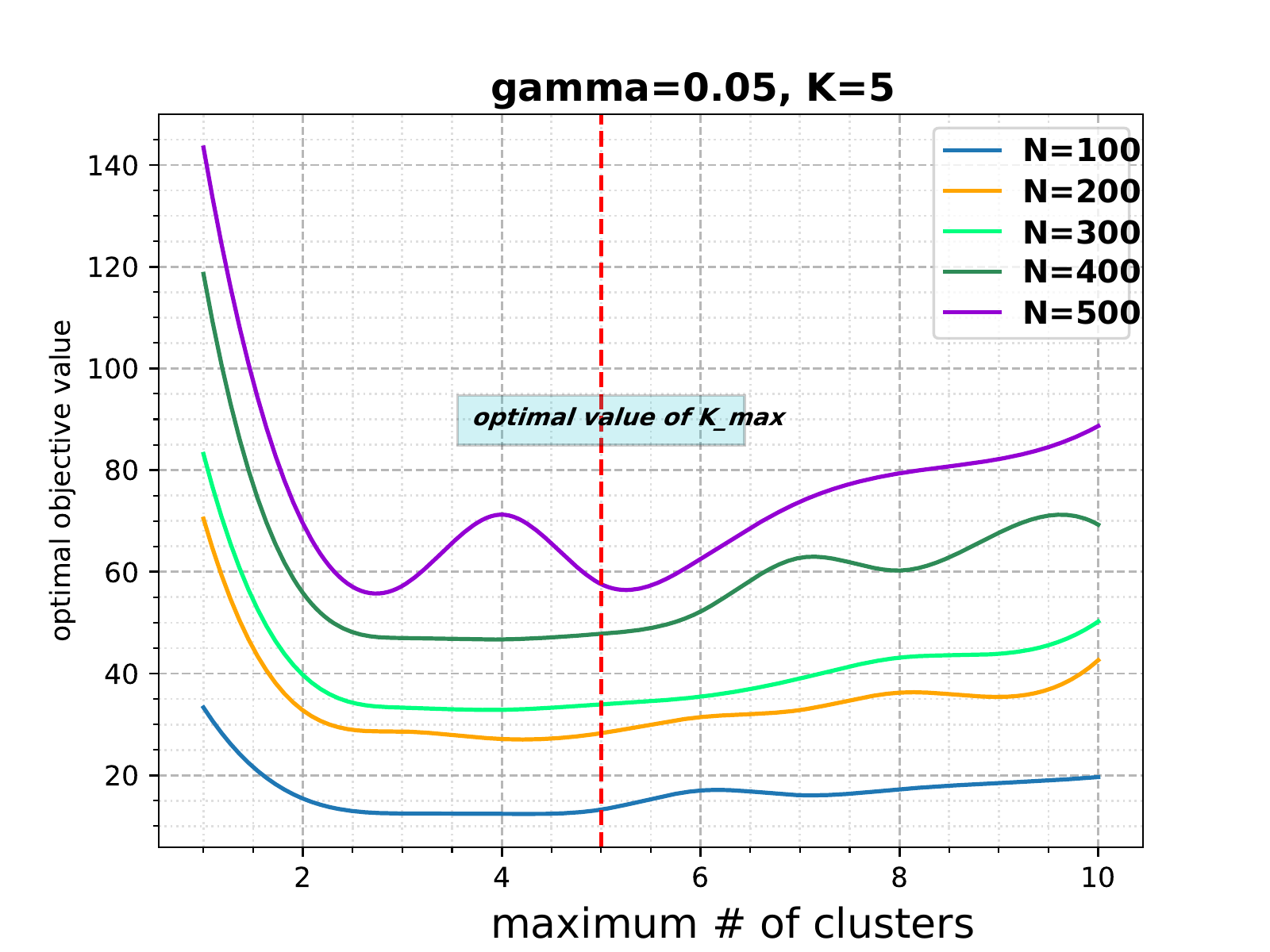} 
&\includegraphics[width=49mm]{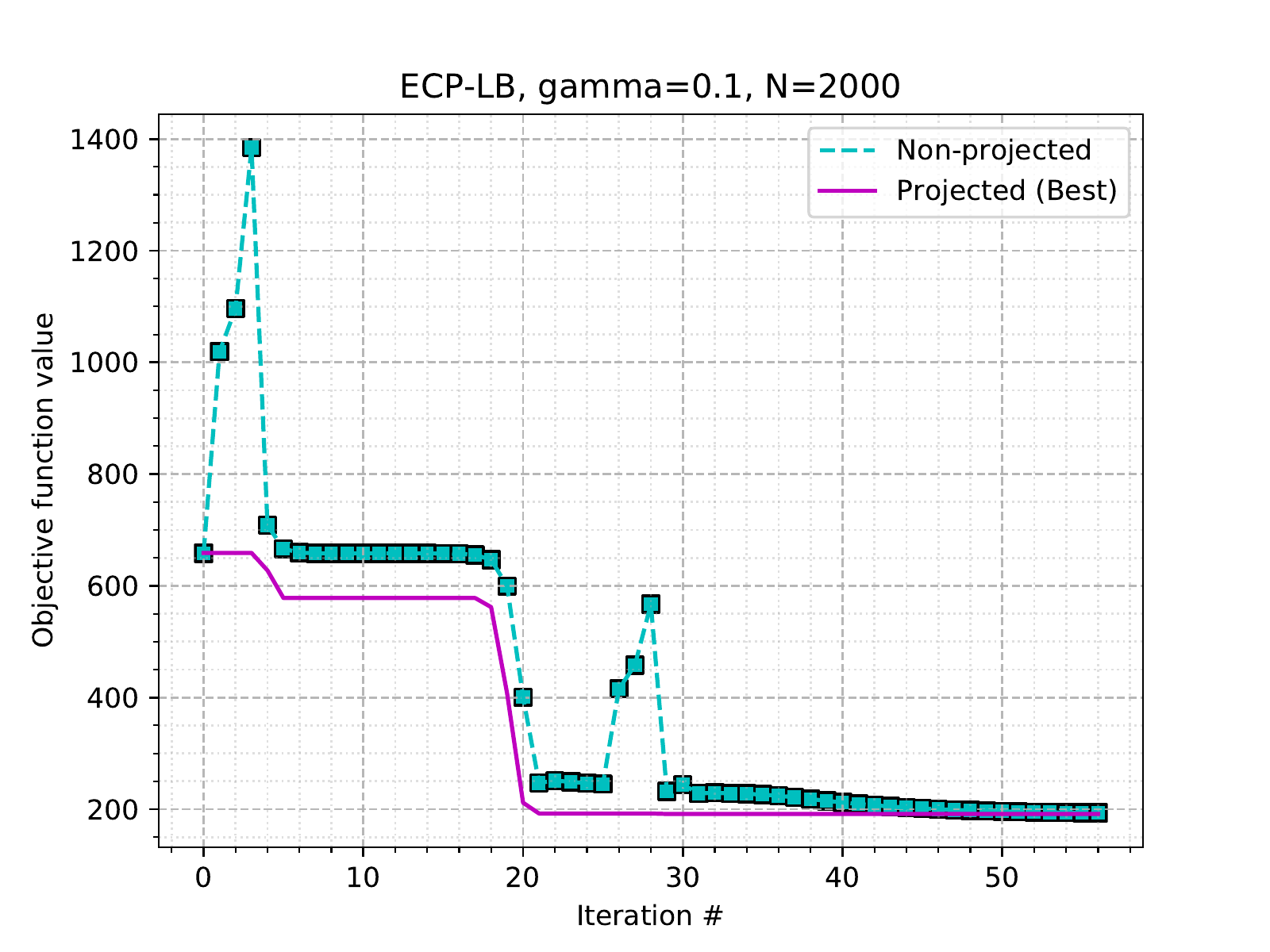}\\
(a) &(b) &(c) &(d)
\end{tabular}
\caption{(a) $\gamma$ vs. optimal objective value, (b) $\gamma$ vs. optimal number of controllers, (c) hyper-parameter $K_{max}$ vs. optimal objective value and (d) Iteration number vs. projected and non-projected solutions objective function values}
\label{fig:LL_tradeoff}
\end{figure*}

Figure \ref{fig:LL_tradeoff} shows the trade-off between different hyper-parameters for ECP-LL algorithm. (a) shows as $\gamma$ increases the optimal objective value also increases and stays relatively constant for very large values of $\gamma$. This is due to the fact that for large $\gamma$, controller placement becomes more packed and at its extreme we would have only one controller to cancel out synchronization cost. Figure \ref{fig:LL_tradeoff} (b) also shows the same pattern that as $\gamma$ increases ECP-LL places fewer controllers in edge network. Figure \ref{fig:LL_tradeoff} (c) shows the optimal value for hyper-parameter $K_{max}$ in ECP-LL algorithm. We validate that the optimal value of $K_{max}$ is the number of inherent clusters in the dataset. Figure \ref{fig:LL_tradeoff} (d) Shows the the values of non-projected and projected~\footnote{At the last step of algorithm we mapped centroids onto the closest edge node available. We call such a solution projected, otherwise we call the solution non-projected.} solutions versus the number of iterations. The projected solution is obtained by setting association probabilities to either zero or one and then projecting the solution centroids onto the data set. We observed that the two converge to the same value across most scenarios. Since the non-projected optimal objective value serves as a lower bound for that of the projected, we can assume ECP-LL has reached, in worst case, a near-optimal solution.

% \usepackage{multirow}
\begin{table*}[htb]
\footnotesize
\caption{Duration and total communication delay as a function of size of dataset $N$ and number of clusters $K$ with $\gamma$ = 0.1. Tuples show completion time (sec), objective value and number of placed controllers triplets. ECP-LL vs BARON}
\label{tab:ll_performance}
\centering
\begin{tabular}{llll}
\toprule
{} &                             N=20 &                               N=40 &                                N=60 \\
\midrule
K=2  &  (0.31,7.10,2), (622.24,15.37,4) &   (0.52,13.24,2), (606.73,37.08,5) &  (0.78,17.50,2), (627.56,585.03,55) \\
K=4  &  (0.48,8.65,7), (614.54,12.93,5) &   (0.77,12.20,7), (610.03,28.41,8) &  (1.08,16.82,7), (624.12,233.88,48) \\
K=6  &  (0.67,7.75,4), (605.20,13.70,6) &   (1.20,15.19,4), (607.32,38.64,9) &   (1.61,23.26,4), (638.92,95.92,13) \\
K=8  &  (0.87,6.53,4), (1630.94,9.39,6) &   (1.34,18.66,4), (606.02,42.28,9) &  (1.96,25.82,4), (618.61,390.73,54) \\
K=10 &  (1.05,6.81,2), (602.03,10.32,5) &  (1.76,12.60,2), (617.59,27.35,10) &   (2.49,19.46,4), (621.64,81.89,15) \\
\bottomrule
\end{tabular}
\end{table*}

\begin{figure*}
\centering
\begin{tabular}{P{4cm}P{4cm}P{4cm}P{4cm}}
\includegraphics[width=49mm]{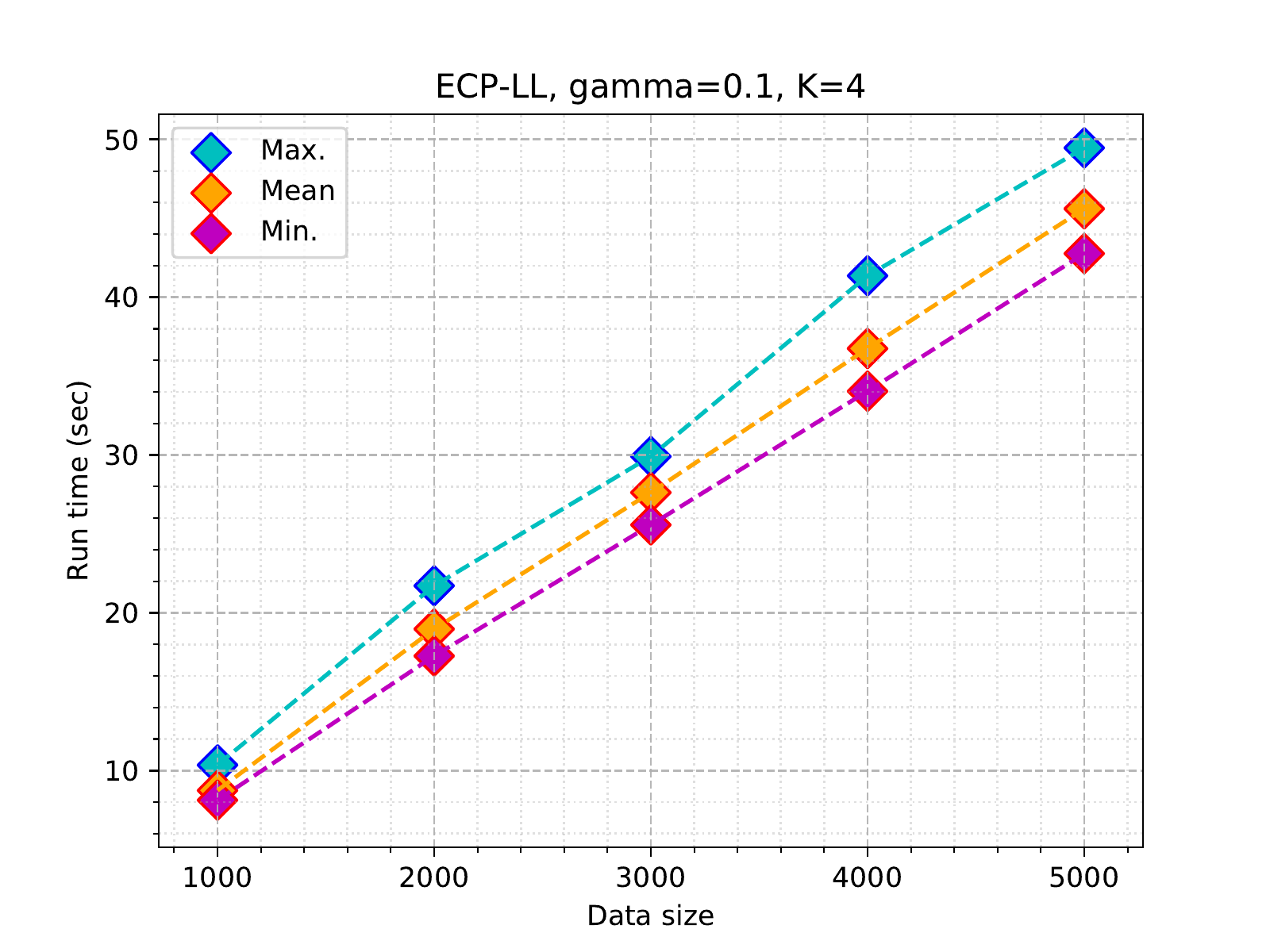}&
\includegraphics[width=49mm]{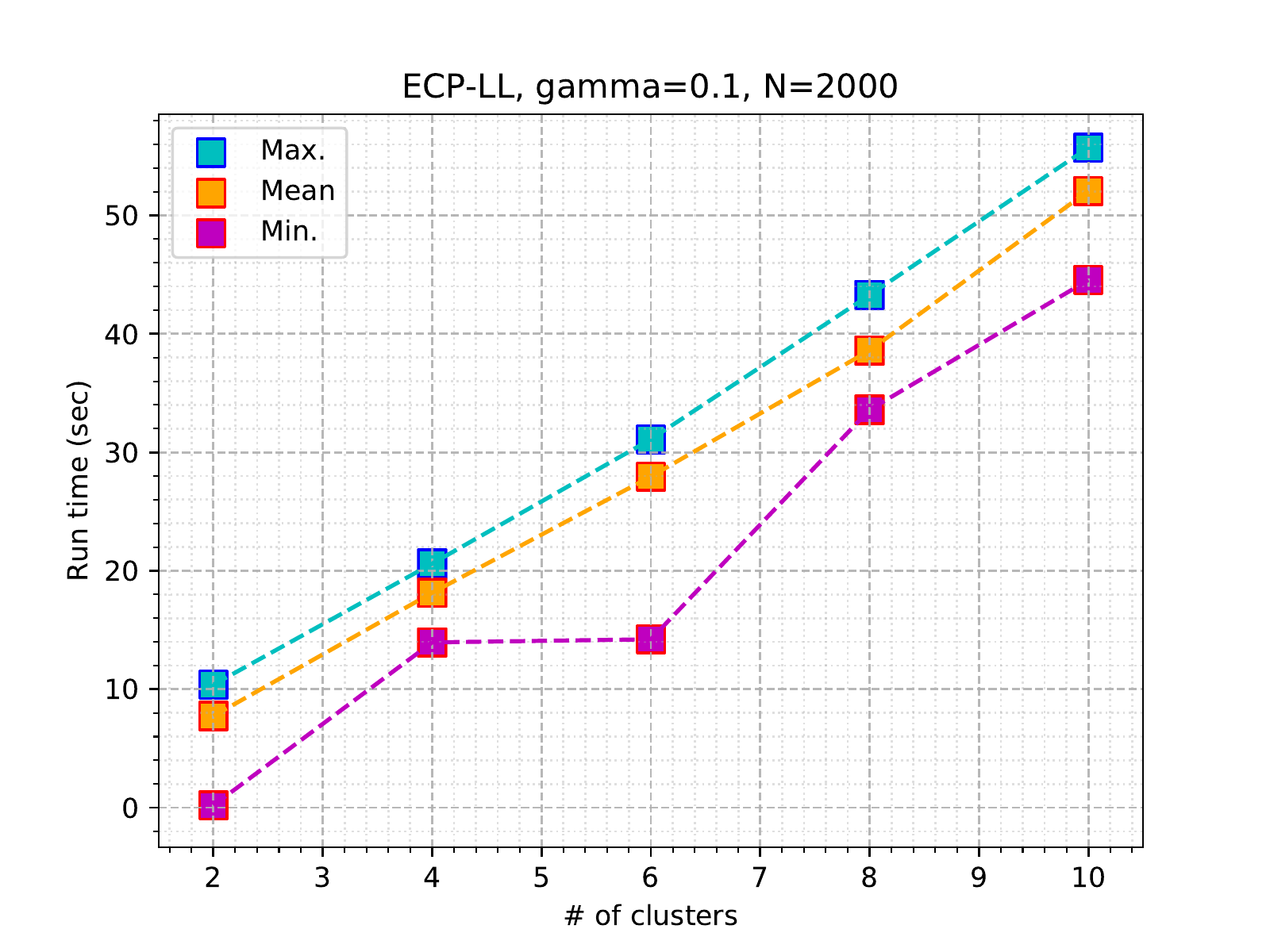}&
\includegraphics[width=49mm]{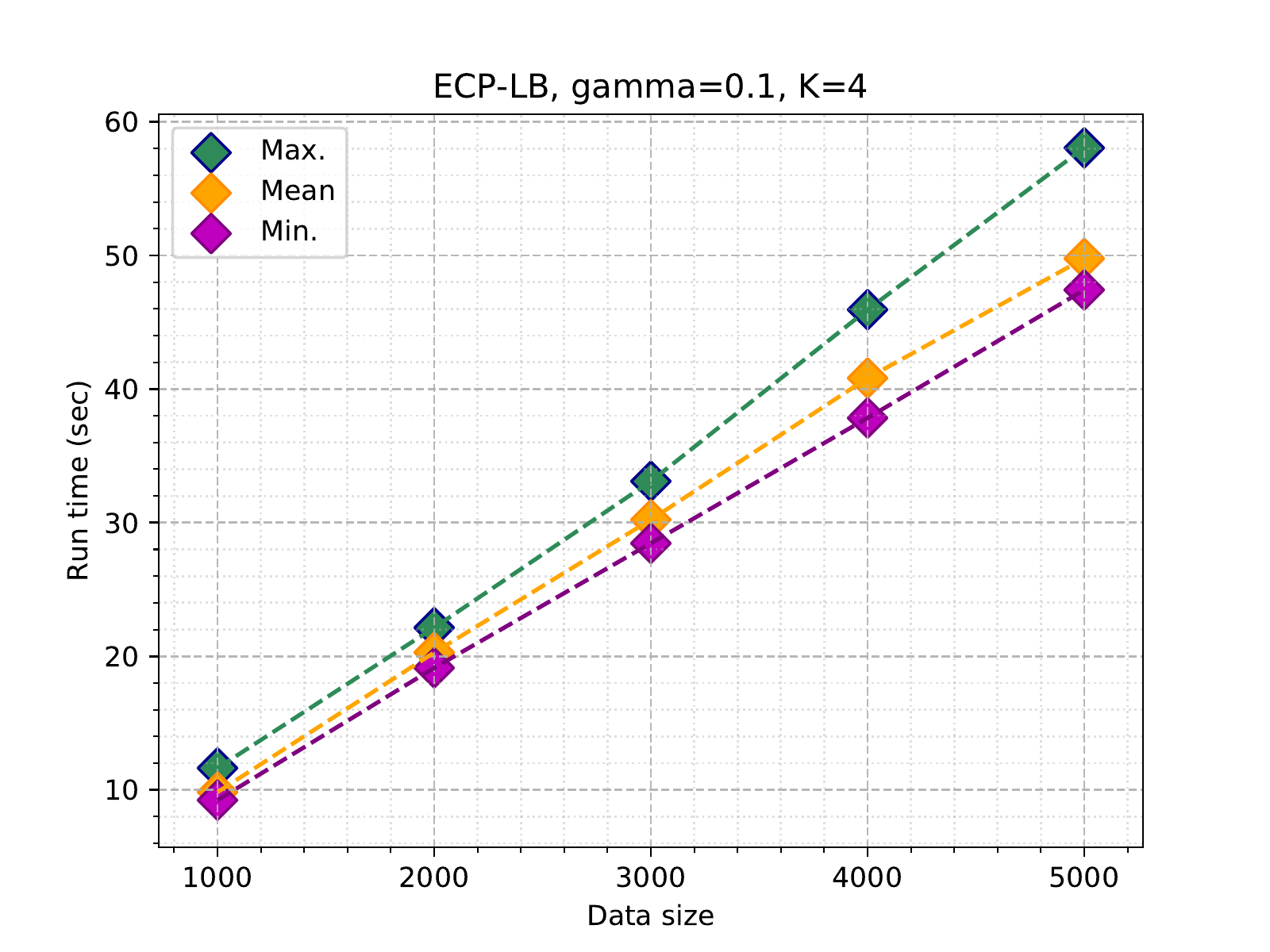}&
\includegraphics[width=49mm]{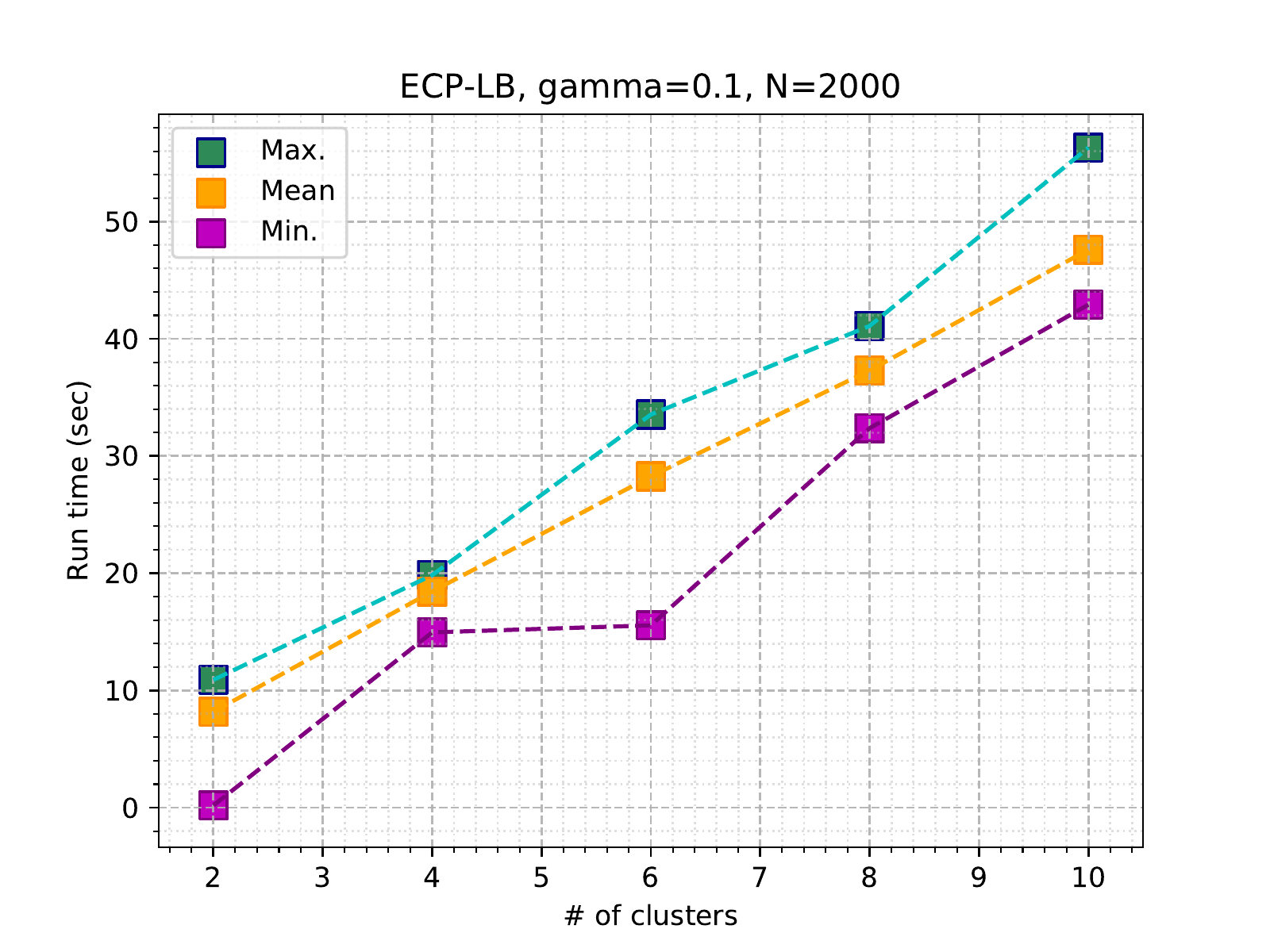}
\end{tabular}
\caption{ECP algorithms run time vs. \# of clusters and data size.}
\label{fig:time_vs_N_clusters}
\end{figure*}

Table \ref{tab:ll_performance} compares performance of ECP-LL against MINLP. While ECP-LL by far outperforms MINLP in terms of total run time, the difference in accuracy is emphasized as problem size increases. ECP-LL and ECP-LB provide consistent performance both in terms of accuracy and speed across different data sizes and varying data clusters as is by design resilient to local minima that riddle the cost function surface. Figure \ref{fig:time_vs_N_clusters} 
illustrates how run time grows linearly as a function of data size and number of clusters.

% \usepackage{multirow}

\begin{table*}[htp]
\footnotesize
\caption{Duration and total communication delay as a function of size of dataset $N$ and number of clusters $K$ and $\gamma$ = 0.1. Tuples show completion time (sec), objective value and number of placed controllers triplets. ECP-LB vs BARON}
\label{tab:lb_performance}
\centering
\begin{tabular}{llll}
\toprule
{} &                            N=20 &                              N=40 &                               N=60 \\
\midrule
K=2  &    (0.93,1.99,5), (2.11,6.21,4) &   (1.63,3.68,5), (373.22,11.88,4) &    (2.36,6.22,2), (906.84,15.24,3) \\
K=4  &   (1.49,2.89,6), (14.87,4.34,5) &    (2.69,4.58,7), (267.35,6.09,6) &     (4.58,6.79,7), (604.13,9.16,6) \\
K=6  &   (2.06,3.52,8), (63.59,4.88,3) &    (3.71,8.14,5), (305.31,8.85,5) &    (5.23,9.10,8), (612.36,12.78,5) \\
K=8  &  (2.81,2.71,11), (16.87,3.78,4) &  (4.61,8.17,11), (458.29,10.56,5) &  (7.06,11.73,11), (604.58,15.07,5) \\
K=10 &   (3.40,3.05,8), (68.99,4.90,4) &    (7.10,5.04,9), (297.49,7.25,4) &  (10.16,7.34,12), (607.64,12.27,5) \\
\bottomrule
\end{tabular}
\end{table*}

\section{Conclusion and Future Work}
\label{sec:conc}
In this work we introduced two multi-objective maximum entropy based clustering algorithms for the problem of Edge Controller Placement in wireless communication networks. ECP-LL and ECP-LB each address a different controller placement topology, and their design is inspired by a Mixed Integer Nonlinear Program. We show that our algorithms outperform state of art MINLP solver, BARON in both speed and accuracy.
Total computational complexity for these algorithms is $O(\tau  N  K_{max}  d)$ which is linear in data size, maximum number of clusters and dimensionality of data.
As an extension to current work we propose a kernelized version of this algorithm. This is significantly important as in most real-world scenarios only the mutual delays between network nodes are provided (e.g. using latency tests) rather than their geospatial coordinates. Another research avenue is to think of a mechanism to project centroids onto the data set at each iteration, as there is no provable guarantee that all centroids will correspond to a data point. This can potentially yield better solutions upon convergence of the algorithm.
Finally one can consider ECP under the assumption that network nodes are mobile, which will require tracking controller design.

\addtolength{\textheight}{-1cm}   % This command serves to balance the column lengths

%%%%%%%%%%%%%%%%%%%%%%%%%%%%%%%%%%%%%%%%%%%%%%%%%%%%%%%%%%%%%%%%%%%%%%%%%%%%%%%%

%%%%%%%%%%%%%%%%%%%%%%%%%%%%%%%%%%%%%%%%%%%%%%%%%%%%%%%%%%%%%%%%%%%%%%%%%%%%%%%%

%%%%%%%%%%%%%%%%%%%%%%%%%%%%%%%%%%%%%%%%%%%%%%%%%%%%%%%%%%%%%%%%%%%%%%%%%%%%%%%%
\section*{APPENDIX}

\textbf{Proof of Proposition \ref{th:has-sol}}

 We can write the coefficient matrix associated with \eqref{eq:LSE} as the block matrix $\Theta \in \Rb^{md\times md}$ with diagonal blocks equal to $\eta I$ and non-diagonal blocks equal to $-\gamma I$ such that $I \in \Rb^{d\times d}$.
%   \begin{equation}
%   \Theta=
%  \begin{bmatrix}
%  &\eta I &-\gamma I &\ldots &
%  -\gamma I\\
%  &-\gamma I &\eta I &\ldots
%  &-\gamma I\\
%  &\vdots &\vdots &\ddots &\vdots\\
%  &-\gamma I &-\gamma I &\ldots &\eta I
%  \end{bmatrix}
%  \end{equation}
 Dividing all rows by constant $-\gamma$ we get $\det(\Theta) = (-\gamma)^{md} \det(\bar{\Theta})$. $\bar{\Theta}$ is a block diagonal matrix with diagonal elements equal to $\alpha I$ and non-diagonal blocks equal to $I$ with $\alpha = -\frac{\eta}{\gamma}$. Using straightforward linear algebra we can transform $\bar{\Theta}$ to an upper triangular matrix:
\small
\begin{align}
\bar{\Theta}
 \times
&\begin{bmatrix}
 & I & 0 &\ldots & 0\\
  &\frac{-1}{\alpha + n -2}I & I &\ldots
 & 0\\\nonumber
 &\vdots &\vdots &\ddots &\vdots\\
  &\frac{-1}{\alpha + n -2}I &\frac{-1}{\alpha + n -3}I &\ldots & I
 \end{bmatrix}
 =\nonumber\\
& \begin{bmatrix}
 &\beta_1 I & \times &\ldots & \times\\
 & 0 &\beta_2 I &\ldots
 & \times\\
 &\vdots &\vdots &\ddots &\vdots\\
 & 0 & 0 &\ldots &\beta_m I
 \end{bmatrix}
 =\Phi
 \end{align}
 \normalsize
 Where $\beta_i = \alpha - \frac{n-i}{\alpha + n -i -1}$ and $\det(\bar{\Theta}) = \det(\Phi) = \prod_{i=1}^m \pa{\beta_i}^d$. We can use simple telescoping to further simplify the product to $\pa{\frac{(\alpha -1)^{m}(\alpha + n -1)}{\alpha + n - (m + 1)}}^d$. This will give $\det \Theta = \pa{\frac{(\gamma m +1)^m\pa{\gamma(n-m) -1)}}{\gamma(n-2m) -1}}^d$ which is well defined for $\gamma \neq \frac{1}{n-m}, \frac{1}{n - 2m}$.

\textbf{Proof of Theorem \ref{thm: theorem2}}

We use the calculus of variations to determine the second order optimality condition for $F$. In particular, consider a given set of optimal centroids $Y=[y_1^T,\hdots,y_m^T]^T\in\mathbb{R}^{md}$ and $Y+\epsilon\Psi$ to be the corresponding set of  perturbed centroids where $\Psi=[\psi_1^T,\hdots,\psi_m^T]^T\in\mathbb{R}^{md}$ is the perturbation vector. The second order condition for optimality states that $\frac{\partial F^2(Y+\epsilon\Psi)}{\partial Y^2}\Big|_{\epsilon=0} > 0$ for all possible perturbations $\Psi$. We obtain:
\begin{align}\label{eq: hessian}
&\frac{\partial F^2(Y+\epsilon\Psi)}{\partial Y^2} = 2\Psi^T\sum_{i=1}^N p(x_i)\Big[\Lambda_i(1+m\gamma)
+ \gamma I\nonumber \\ & - 2\gamma\Gamma_i^TE-2\frac{1}{T}\Theta_i \Big]\Psi + 4\beta\Psi^T\Big(\sum_{i=1}^N p_i \mathcal{L}_i^T\mathcal{L}_i\Big)\Psi,
\end{align}
where $\Lambda_i=\text{diag}\big(p(y_1|x_i)~p(y_2|x_i)~\hdots~p(y_m|x_i)\big)$, $m$ is the number of centroids, $E = 1^T_m\otimes I_d$, $1_m\in\mathbb{R}^m$ is a vector of $1$'s, $\Gamma_i=\sum_{j=1}^mp(y_j|x_i)E_j \Psi$, $E_j=e_j^T\otimes I_d$, $e_j$ is the basis vector in $\mathbb{R}^m$ with $j$-th entry as $1$, $\Theta_i=\sum_{j=1}^m p(y_j|x_i)T_{ji}^TT_{ji}$, $T_{ji}=(y_j-x_i)^TE_j +\gamma\sum_{k=1}^m (y_j-y_k)^T(E_j-E_k)$, and $\mathcal{L}_i=\sum_{j=1}^m p(y_j|x_i)T_{ji}$. We claim that the Hessian in (\ref{eq: hessian}) is positive for all perturbations $\Psi$ if and only if the first term in (\ref{eq: hessian}) is positive. Clearly, when the first term is positive this is true owing to the non-negativity of the second term ($(\mathcal{L}_i\Psi)^T(\mathcal{L}_i\Psi)\geq 0$). This establishes the {\em if} part of our claim. For the {\em only if} part we show that there exists a non trivial perturbation when the matrix in the first term is not positive definite for which the second term becomes zero. In fact, let $y_{j_0}$ be the centroid that splits into two further centroids. Let $\psi_k=0~\forall ~k \neq j_0$ and $\psi_{j_0}$ such that $p(y_{j_0}|x_i)(y_{j_0}-x_i)^T\psi_{j_0}=0$ (a non trivial $\psi_{j_0}$ exists); this choice of perturbation results in the second term being zero. Thus, whenever the matrix in the first term loses rank we can construct a perturbation that makes the second term zero. This is the phase transition condition in (\ref{eq: phaseTran}).

%\section*{ACKNOWLEDGMENT}

%%%%%%%%%%%%%%%%%%%%%%%%%%%%%%%%%%%%%%%%%%%%%%%%%%%%%%%%%%%%%%%%%%%%%%%%%%%%%%%%

%\nocite{Parekh2015DeterministicAspects}
\bibliographystyle{apalike}
\bibliography{references}
\end{document}